\documentclass[letterpaper]{article} 

\usepackage{aaai25}  
\usepackage{times}  
\usepackage{helvet}  
\usepackage{courier}  
\usepackage[hyphens]{url}  
\usepackage{graphicx} 
\usepackage{natbib}  
\usepackage{caption} 
\usepackage{algorithm}
\usepackage{algorithmic}

\usepackage{newfloat}
\usepackage{listings}
\DeclareCaptionStyle{ruled}{labelfont=normalfont,labelsep=colon,strut=off} 
\floatstyle{ruled}
\newfloat{listing}{tb}{lst}{}
\floatname{listing}{Listing}
\nocopyright

\title{Retrieval-Augmented Generation by Evidence Retroactivity in LLMs}
\author{
    Liang Xiao\textsuperscript{\rm 1}\thanks{This work was completed during an internship at Xiaomi Corporation.},
    Wen Dai\textsuperscript{\rm 2},
   Shuai Chen\textsuperscript{\rm 2},
  Bin Qin\textsuperscript{\rm 2},
   Chongyang Shi\textsuperscript{\rm 1}\thanks{*Corresponding authors},
Haopeng Jing\textsuperscript{\rm 1}, 
 Tianyu Guo\textsuperscript{\rm 1}\\
}
\affiliations{
    \textsuperscript{\rm 1}Beijing Institute of Technology, \textsuperscript{\rm 2}Xiaomi Corporation

    \{patrickxiao, cy\_shi, jinghp, guotianyu\}@bit.edu.cn, \{daiwen, chenshuai3, qinbin\}@xiaomi.com
}

\usepackage{bibentry}
\usepackage{amsmath}
\usepackage{multirow}
\usepackage{booktabs}
\usepackage{caption}
\usepackage{subfigure}
\usepackage{graphicx}

\begin{document}

\maketitle

\begin{abstract}

Retrieval-augmented generation has gained significant attention due to its ability to integrate relevant external knowledge, enhancing the accuracy and reliability of the LLMs' responses. Most of the existing methods apply a dynamic multiple retrieval-generating process, to address multi-hop complex questions by decomposing them into sub-problems. However, these methods rely on an unidirectional forward reasoning paradigm, where errors from insufficient reasoning steps or inherent flaws in current retrieval systems are irreversible, potentially derailing the entire reasoning chain. For the first time, this work introduces $\textbf{Retro}$active $\textbf{R}$etrieval-$\textbf{A}$ugmented $\textbf{G}$eneration (RetroRAG), a novel framework to build a retroactive reasoning paradigm. RetroRAG revises and updates the evidence, redirecting the reasoning chain to the correct direction. RetroRAG constructs an evidence-collation-discovery framework to search, generate, and refine credible evidence. It synthesizes inferential evidence related to the key entities in the question from the existing source knowledge and formulates search queries to uncover additional information. As new evidence is found, RetroRAG continually updates and organizes this information, enhancing its ability to locate further necessary evidence. Paired with an Answerer to generate and evaluate outputs, RetroRAG is capable of refining its reasoning process iteratively until a reliable answer is obtained. Empirical evaluations show that RetroRAG significantly outperforms existing methods. 

\end{abstract}

%

\section{Introduction}
Large language models (LLMs), such as ChatGPT\cite{chatgpt} and ChatGLM\cite{chatglm4}, have demonstrated outstanding performance across a wide range of natural language processing tasks. However, despite the vast amount of knowledge stored during training, these models still exhibit a tendency to generate hallucinatory content, resulting in unverified or factually incorrect answers\cite{Hallusurvey,ClashEval}. To address this issue, the Retrieval-Augmented Generation (RAG) framework is leveraged to acquire and subsequently inject relevant external source knowledge into the LLM's prompt, significantly enhancing the accuracy and reliability of LLM's responses\cite{rag_ori,rag,AAAI24RAG}. 
\begin{figure}[!t]
    \centering
  \includegraphics[width=\linewidth]{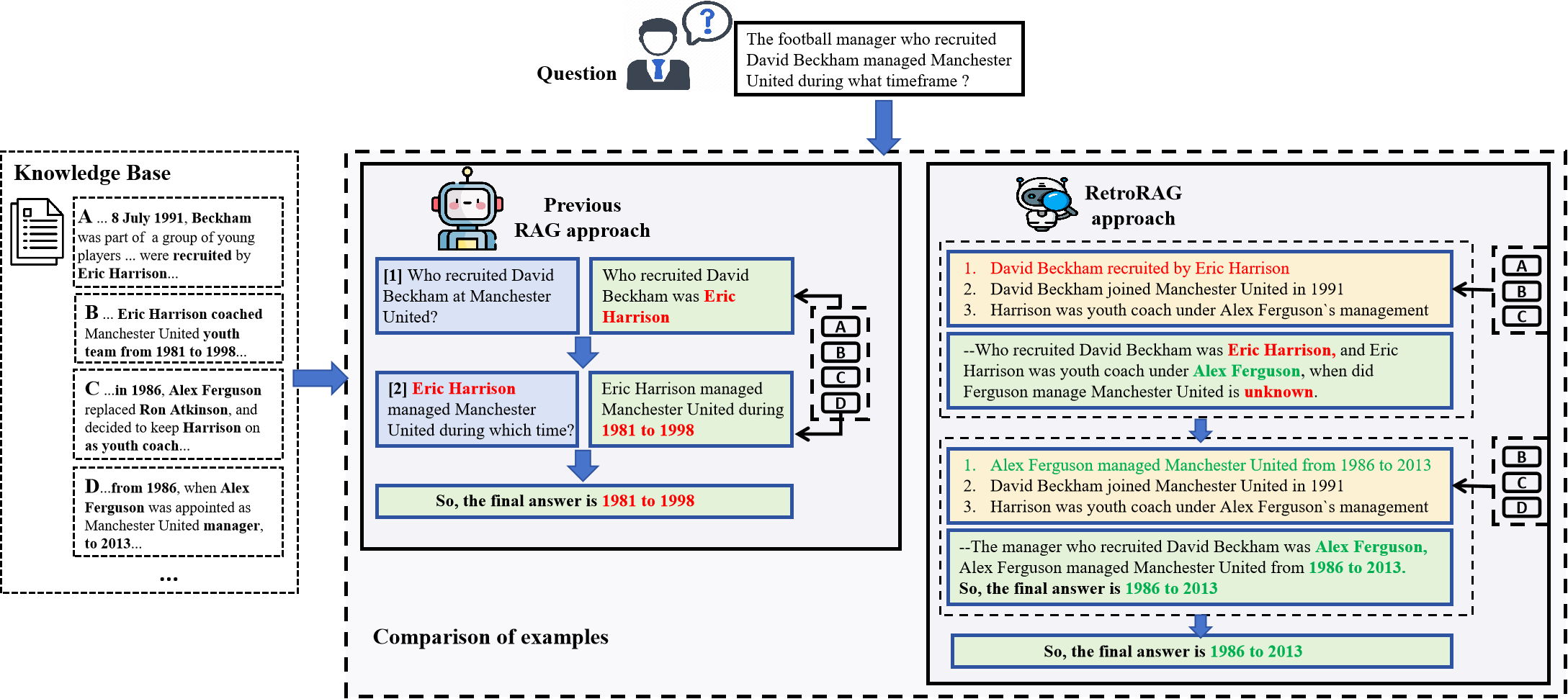}
  \vspace{-7mm}
  \caption{An example of previous RAG approaches causes hallucinatory content due to their unidirectional forwards reasoning paradigm, and how RetroRAG address this issue.}
\label{fig:motivation}
\vspace{-6mm}
\end{figure}
\par Traditional retrieval-augmented models typically retrieve and extract knowledge documents once based on the initial query, these approaches struggle with addressing multi-hop complex questions due to insufficient knowledge. To tackle this issue, recent studies have transformed the single retrieval-generating into a dynamic multiple retrieval-generating process. These approaches decompose the complex question into several sub-questions, and obtain the final output by answering all these sub-questions\cite{emnlp-enhancing,ircot,selfask,react}. Even though the latest approaches have been proposed to improve the effectiveness of knowledge documents retrieval\cite{asai2024selfrag,search-in-chain}, current RAG frameworks are susceptible to the threat of insufficient reasoning from the documents which are factual, and related, but irrelevant due to the inherent flaws of current retrieval systems\cite{influence_too_much}, and cause external hallucination. As illustrated in Figure 1,  the excessive focus on the local sub-question of \textit{who recruited Beckham?} obtaining the incorrect answer of \textit{Eric Harrison}, the youth coach of Manchester United who \textit{recruited Beckham} in the youth team but never as the \textit{manager}, rather than the manager at time \textit{Alex Ferguson}, and mislead the following reasoning steps. 
\par We argue that this flaw originates from the \textbf{Unidirectional Forwards} reasoning paradigm inherent in traditional RAG methods. In this paradigm, any errors produced during reasoning steps are irreversible for the whole reasoning chain. Although the paradigm can be altered by enabling LLMs to continuously reason from scratch through the cumulative retrieval of documents\cite{ircot,metarag}, due to the limited useful information in documents, excessive retrieval would introduce much more noise information which distracts LLMs to engage in over-reasoning and build erroneous correlation, thereby generates hallucinatory content\cite{yu2023chainofnote,over_reasoning,influence_too_much,RA-ISF}. 
\par To address these issues, we refer the investigative process of the detective, that iteratively collates evidence to gather all related factual information, and through evidence discovery process to validate the relevance of evidence then update them, while uncovering unresolved issues along the way, to ultimately reach a valid conclusion. By continuously revising and reconstructing the reasoning chain through the evidence collation and discovery process, a comprehensive and definitive conclusion can be obtained\cite{evidencediscover,detective}. This evidence-collation-and-discovery structure allows LLMs to rethink and revise the reasoning chain through a \textbf{Retroactive} paradigm, correcting  previous errors resulting from insufficient information by utilizing newly discovered evidence. As illustrated in Figure 1, LLMs can collate the evidence that \textit{Eric Harrison} was the youth coach of the manager \textit{Alex Ferguson} since \textit{1986}, and \textit{Ferguson} was actually the \textit{manager}, then update the reasoning process and answer correctly. Inspired by this detective-like approach, we introduce the \textbf{Retro}active \textbf{R}etrieval-\textbf{A}ugmented \textbf{G}eneration framework (RetroRAG).
\par To effectively generate and utilize evidence, there are two aspects need to be considered: (1)\textit{The Effectiveness}: the evidence should align with external inherent knowledge\cite{evidence_gather}, while being attributable\cite{RARR} and relevant\cite{RA-ISF}, to avoid the irrelevant noise; (2)\textit{Dynamic Updating}: the evidence should be continuously updated based on newly discovered information and overarching question, rather than being confined to local sub-questions. Hence, RetroRAG constructs \textbf{E}vidence-co\textbf{LL}ation-and-discov\textbf{ERY} framework (\textbf{ELLERY}) to retrieve, generate, and update the evidence, which involves two major components: (1)\textbf{Evidence Collation} retrieves relevant documents from the retrieval corpus as the \textbf{source evidence}, which would be utilized as doubtless material to generate \textbf{inferential evidence}, besides serving as the primary reference for the answering.  The source evidence will be continuously updated as the question-solving process progresses, to mitigate the influence of retrieved irrelevant information; (2)\textbf{Evidence Discovery} first generates as much inferential evidence related to the key entities in question as possible from the source evidence. Then, inferential evidence will be filtered from the perspectives of relevance to the question and its attribution to the source evidence, to ensure the effectiveness. Inferential evidence would also be updated to only remain the most relevant parts. While both evidence would be used to help generate answers, the gap between the stored evidence and the initial question would be simultaneously analyzed, and the search-query would be proposed for retrieving more information in need.
\par Along with the \textbf{Answerer} to generate and evaluate the answer, RetroRAG provides an approach for refining effective reasoning chains through credible evidence. The experimental results on two multi-hop question answering (QA) datasets verify the effectiveness and state-of-the-art performance of RetroRAG, while also demonstrating the explainability in the reasoning process. 
\par The contributions of this paper are summarized as:  
\begin{itemize}
    \item We introduce RetroRAG approach, an innovative retrieval-augmented generation framework. Unlike existing RAG approaches uses an unidirectional forwards reasoning paradigm that cannot reverse the error in preceding reasoning steps,  RetroRAG uses a retroactive reasoning paradigm that can revise and reconstruct the reasoning chain through two types of evidence, provides effective answers with less hallucination.
    \item     To the best of our knowledge, this is the first time an evidence-collation-and-discovery framework has been proposed and used in a retrieval-augmented framework, which generates and updates the evidence to support the reasoning process, significantly enhances knowledge retrieval performance on question-answering tasks.
\end{itemize}
\section{RELATED WORK}
\subsection{Hallucination in Large Language Model}
Currently, hallucination is referred as generated content that either does not align with real-world facts or deviates from the source material and self-consistency\cite{huang2023survey,ye2023cognitive}. In the context of question-answering tasks, hallucination specifically manifests as the generation of arbitrary, and incorrect answers. This phenomenon occurs because, in cases of hallucination, the internal consistency of the generation process in LLMs is unstable. \cite{selfcheckgpt,selfcontradictoryhallucinationslargelanguage,farquhar2024detecting}.
Some studies consider addressing the hallucination problem based on the tendencies of generation from LLMs, they generate multiple outputs and then employ a majority voting strategy to obtain relatively reliable answers\cite{wang2022self,huang2022large}. More studies consider that the inconsistency generation of LLMs stems from the lack of knowledge, therefore, they introduce reliable external knowledge through the Retrieval-Augmented Generation (RAG) framework, to enhance the factual or specific domain knowledge of LLMs\cite{he2022rethinking,RARR,siriwardhana-etal-2023-improving,ram2023context_rag}. Besides, some methods enhance the LLMs to better perceive factual information by fine-tuning the model with the external knowledge \cite{lee2022factuality,tian2023fine}.  
\subsection{Retrieval Augmented Language Model}
Many studies have demonstrated the impressive performance of the retrieval-augmented language model (RALM) in various natural language tasks, which is enhanced by the provision of detailed and specific external knowledge to supplement LLMs\cite{rag_ori,RAG_pre,replug}. These models typically employ a retriever to obtain a set of relevant documents from a knowledge corpus, such as Wikipedia, to enhance the effectiveness of the answers. While initial RALM performs the single-time retrieval strategy, which extracts knowledge once based on the user`s initial query\cite{he-etal-2021-efficient,izacard-grave-2021-leveraging,ram2023context_rag}, recent studies have focused on multi-time retrieval models to overcome the issues of insufficient knowledge, due to retriever may focus only on parts of the query when addressing multi-hop complex questions. Some models decompose the initial query into multiple sub-questions, then iteratively retrieve knowledge and answer these sub-questions, until the original query can be finally resolved\cite{react,selfask,emnlp-enhancing,search-in-chain}; while the others construct an iterative process of holistic thinking, continuously increasing the amount of knowledge retrieved based on unsolved questions, until effective reasoning can be achieved\cite{ircot,metarag}, and more recent studies have explored fine-tuning certain components to enhance the reliability of retrieval process\cite{yan2024corrective,RA-ISF}.  
\par All these approaches are proven to be effective. However, due to factually related irrelevant documents from the inherent flaws of the current retrieval system, they address issues of insufficient reasoning and over-reasoning, respectively. This paper addresses the aforementioned issues by exploring the application of evidence, to construct a retroactive reasoning process. Through continuously generating and updating credible evidence, our work builds an effective RAG framework to address the hallucination in question answering task without fine-tuning or pre-training of LLMs.

\section{Methodology}
Existing retrieval augmented methods, due to their unidirectional forward reasoning paradigm, are prone to the risk of external hallucination from factually related irrelevant documents. Since the process of answering decomposed sub-questions can be equivalently regarded as the process of obtaining sub-evidence to answer the initial question, as illustrated in Figure 2, previous approaches have employed a linear paradigm of progressive sub-evidence generation, where the generation of each node is highly depends on the previous nodes. Although the verification of knowledge can prevent the emergence of subsequent unreliable node \textcircled{B}, it is incapable of correcting erroneous validation node \textcircled{C} caused from the inherent flaws of current retrieval systems, like \textit{Eric Harrison} in Figure 1 when it has reached node \textcircled{D}. The LLMs will propagate the erroneous information as definitive knowledge, leading to inaccuracies in the following output. 
\par By generating and updating of evidence, RetroRAG enables LLMs to refine their knowledge by integrating newly evidence \textcircled{D}, \textcircled{J}, \textcircled{K}, with the previous evidence \textcircled{A}, \textcircled{B}, \textcircled{C} based on the relevance to the question, retaining only the most pertinent evidence. This process allows for a more comprehensive understanding as new evidence is integrated(for instance, \textit{Eric Harrison is the youth coach of Alex Ferguson}),  while erroneous nodes at any stage are discarded (for instance, \textit{Harrison recruited Beckham as manager} will be correct). The correct nodes \textcircled{A}, \textcircled{J}, \textcircled{K} will be preserved to the next stage. Essentially, it can be considered that RetroRAG constructs a graph-based thinking structure.

\subsection{Overview}
We propose Retroactive Retrieval-Augmented Generation (RetroRAG) framework to tackle the issue of insufficient reasoning and over-reasoning,  which applies Answerer to generate and evaluate credible answers, and Evidence-coLLation-and-discovERY (ELLERY) framework retrieves, generates, and updates the evidence.  
\par In the QA scenario, the target of RetroRAG is to generate an answer $a$ to the given question $q$ with key entities $k$. As illustrated in Figure 3, RetroRAG utilizes the iteration application of two processes: (1) \textit{Answerer} generates an answer based on the current knowledge context, and determines if a consistent response can be generated within the current knowledge context. (2) \textit{ELLERY} obtains documents $D_{Q}$ from the retrieval corpus $D=\{d_{i}\}_{i=1}^{|D|}$(with Wikipedia dumps served as the primary data source in this study) related to the question, as the source evidence, and by which generates credible evidence $E$, and re-queries $q_{r}$ based on $q$, $k$, $D_{Q}$, and the last reasoning chain $r$. ELLERY continues this process of collating and discovering evidence until a definitive answer is obtained by Answerer. The details of the prompts we designed and used will be introduced in Section A of the Appendix.

\subsection{Answerer}

The main target of Answerer is to generate a reliable answer to the question. To achieve this, the Answerer first generates pseudo-answer $a$ and corresponding reason $r$ with the current quantities of information $E$ through an answer-generator, which utilizes COT-prompt $M_{C}$ with a low-temperature parameter to obtain a more fixed output. 
\par To assess whether the current knowledge context is sufficient, we refer Self-Consistency (SC) concept that outputs of LLMs should converge towards the correct answer under a strong knowledge context, with which LLMs are capable of constructing reliable reasoning and answers. To this end, we employ an SC-generator with a direct-answering-prompt $M_{D}$ and high-temperature parameter to obtain monitoring answer $a_{sc}$ with more divergent thinking pattern and the same knowledge content of $a$, and the similarity between these two outputs can assess the self-consistency score and the degree of hallucination. We designed an LLM-based evaluator $S$ to convert scores, e.g., similarity or relevance, into the probability of generating indicative tokens (e.g., 'yes' or 'no'). We calculate the similarity scores $s_{sc}$ through LLM-based evaluator $S_{sc}$ as the following formulation: 
\begin{figure}[!t]
    \centering
  \includegraphics[width=\linewidth]{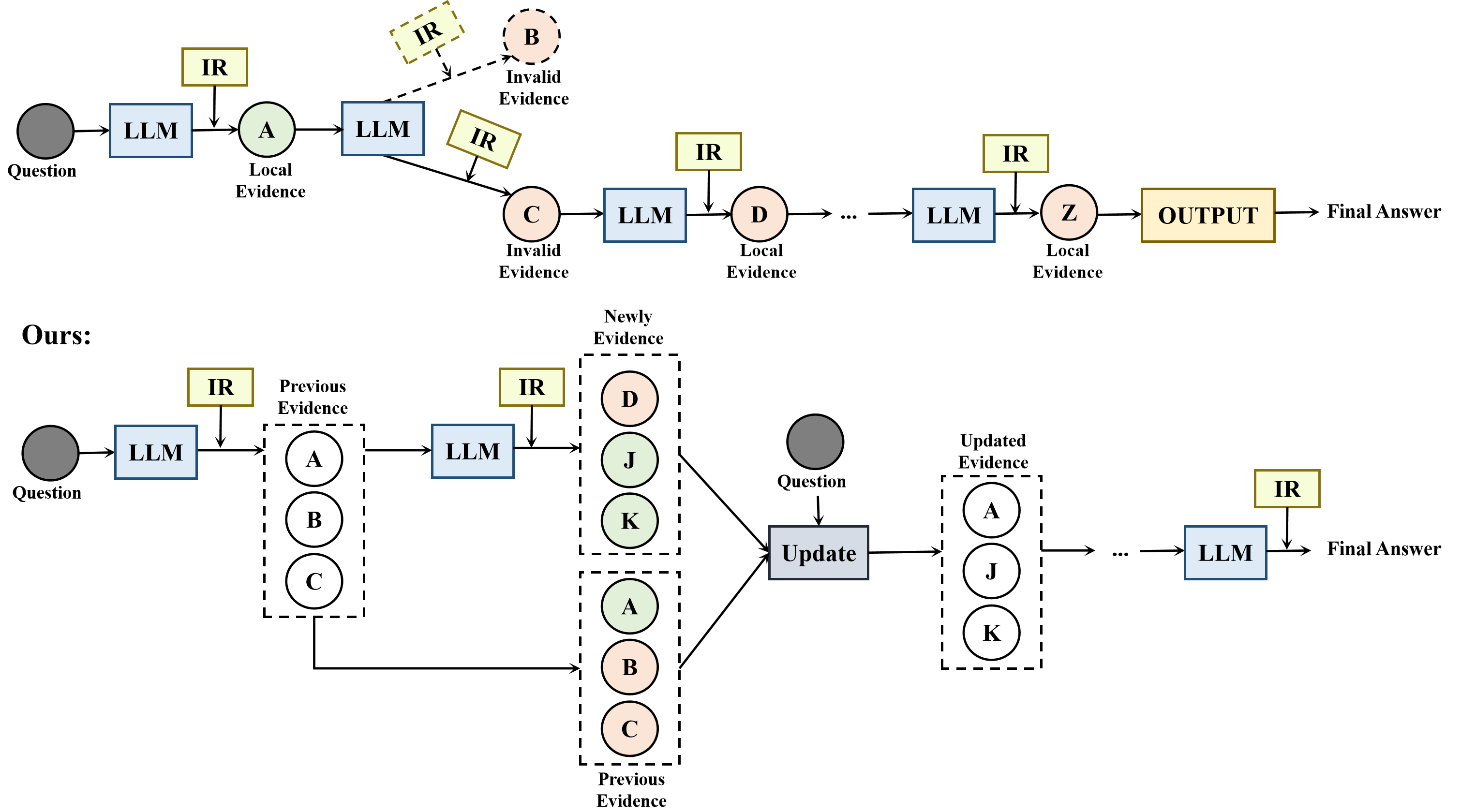}
  \vspace{-7mm}
  \caption{Comparison between previous methods and RetroRAG in mechanism.}
\label{fig:Comparison}
\vspace{-6mm}
\end{figure}

\vspace{-3mm}
\begin{figure*}[!t]
  \includegraphics[width=2\columnwidth]{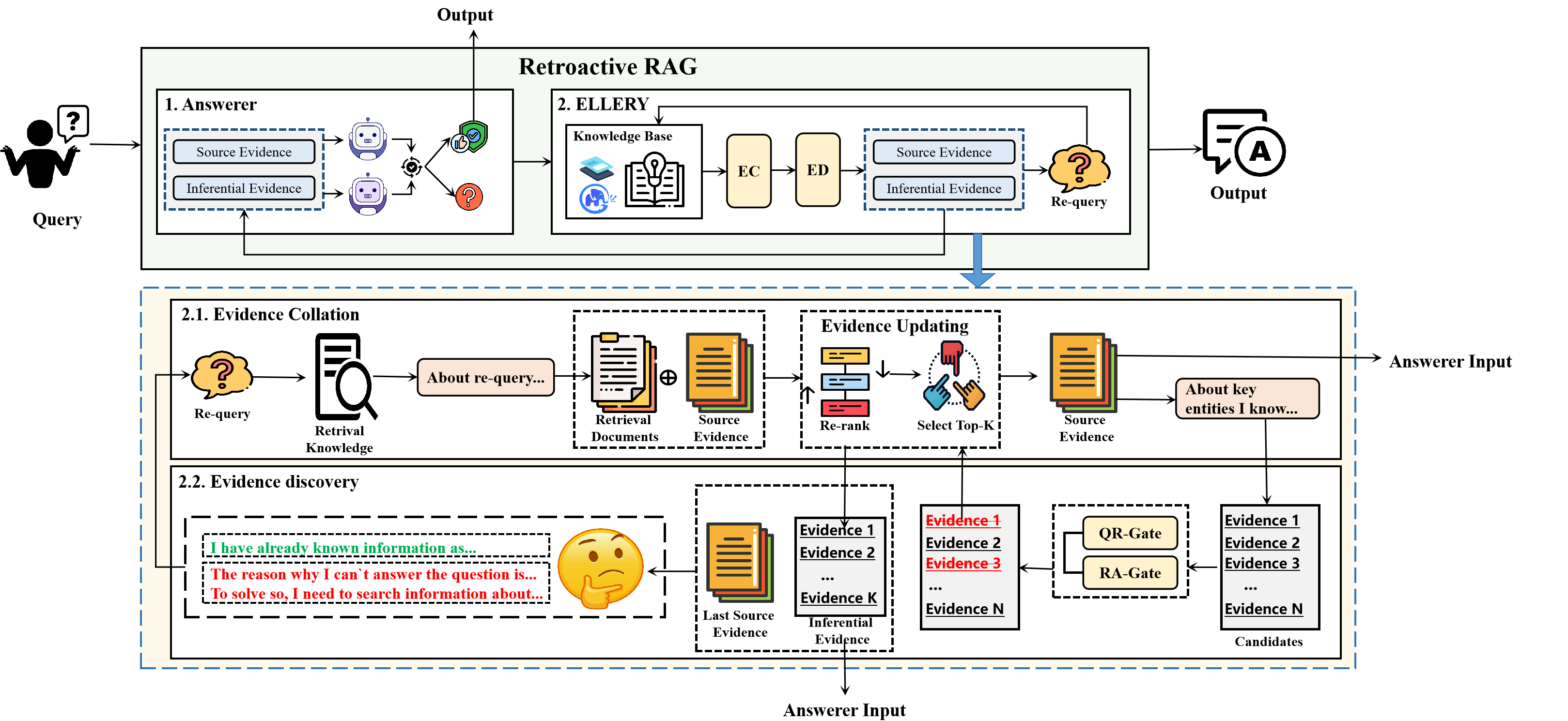}
  \vspace{-3mm}
  \caption{Overview of our RetroRAG structure.}
  \label{modelmain}
\vspace{-4mm}
\end{figure*}

\begin{equation}
\begin{aligned}
     x &= LLM([a,a_{sc},q],M_{sc}) \\
  S_{sc} &=\frac{P(x='yes'|([a,a_{sc},q],M_{sc}))}{\sum_{i \in ['yes','no']}P(x=i|([a,a_{c},q],M_{sc}))}\\
      s_{sc} &=S_{sc}(([a,a_{sc},q],M_{sc}))\\
\end{aligned}
\end{equation}
Where $M_{sc}$ is a customized prompt, and a threshold \textit{t} is utilized to govern the model`s output, only when $s_{sc}>t$, the iteration process is stop, and output the answer as the final answer. A higher value of \textit{t} implies that a more stringent requirement of knowledge context. And follow the research from \cite{metarag}, a declarative assessor is implemented to ensure the standardization of the answer.

\subsection{Evidence Collation and Discovery}
In each round of iteration, the initial query $q$ and its key entities $k$ will be fixed as constant input for ELLERY structure. The ELLERY structure obtains and updates source evidence through Evidence Collation, while based on which generating inferential evidence and proposing re-querie to obtain the required knowledge through Evidence Discovery. 
\subsubsection{Evidence Collation}
In the \textit{L-}th iteration, we use the search query generated from end of the last round $q_{s}^{(L-1)}$,which is designed to specifically target missing information, to retrieve relevant passages $D_{C}^{(L)}$. And concatenate $q_{s}^{(L-1)}$ with $q$ to obtain the matching query $q_{m}^{(L)}=[q,q_{s}^{(L-1)}]$, which is designed to match the most relevant evidence within the current knowledge context, while avoiding the issue of deviating from the initial question by focusing too much on the generated search query. Specifically, we set $q_{s}^{(0)}=q_{m}^{(0)}=q$ as the initialization. After obtaining $D_{C}^{(L)}$, we merge them with the last stored source evidence $E_{s}^{(L-1)}$, and apply the customized prompt  $M_{e}$ with LLM to individually score each source evidence candidate, ranging from 0 to 1, to assess the contribution of source evidence to answering the question. Since source evidence should be updated to ensure the progression of current answering process, the scoring process of evaluators $S_{E_{s}}$ can be formulated as:
 \begin{equation}
\begin{aligned}
  s_{E_{s}} &=S_{E_{s}}([E_{s}^{(L-1)}\cup D_{C}^{(L)},q_{m}^{(L)}],M_{e}))\\
\end{aligned}
\end{equation}
Based on  $s_{E_{s}}$, we can extract the top-N source evidence as the current source evidence $E_{s}^{(L)}$, which would be sent to Answerer to help the answering process in the \textit{L-th} round, and be used to generate the inferential evidence and the re-query in the failure answering case. 

\subsubsection{Evidence Discovery}
Follow the actual detective, we uses \textbf{Deductive} method to generates inferential evidence, which contains two steps: (1)\textit{Deductive Reasoning}: we utilize an LLM-based evidence generating prompt $M_{IE}$ to generate inferential evidence candidates $e_{ic}^{(L)}$ related to $k$ from $E_{s}^{(L)}$, to obtain as much \textbf{deductive inference} to the initial question as possible, from the current retrieval documents. (2)\textit{Hypothesis Testing}: we design two specified LLM-based gated prompt $M_{qr}$ and $M_{ra}$ to calculate the Question-Relevance (QR) score and the Reference-Attribution (RA) score with their LLM-based evaluator $S_{qr}$ and $S_{ra}$ , to ensure the effectiveness of each inference. The specific functions of these two evaluation scores are as follow:
\begin{itemize}
    \item \textit{Question-Relevance (QR)}: On knowing the last inferential evidence $E_{i}^{(L-1)}$, if $e_{ic}^{(L)}$ could be directly related to answering the matching query $q_{m}^{(L)}$:
    \item \textit{Reference-Attribution (RA)}: If the claim of $e_{ic}^{(L)}$ can be directly found in any claims of $E_{s}^{(L)}$.
\end{itemize}
Through these step, we only reserve the useful and confirmed inference as evidence, and this process can be formulated as:
 \begin{equation}
\begin{aligned}
  s_{qr} &=S_{qr}([e_{ic}^{(L)},E_{i}^{(L-1)},q_{m}^{(L)}],M_{qr}))\\
    s_{ra} &=S_{ra}([e_{ic}^{(L)},E_{s}^{(L)}],M_{ra})\\
e_{ic}^{(L)}&=((s_{qr}>0.5)\cap(s_{ra}>0.5))e_{ic}^{(L)}
\end{aligned}
\end{equation}

Next, we merge $e_{ic}^{(L)}$ and last inferential evidence $E_{i}^{(L-1)}$. Following the updating process of source evidence, we apply the same evaluator $S_{e}$  score each inferential evidence candidate. Since inferential evidence should align with the initial question for a long term memory, the scoring process of evaluators $S_{E_{i}}$ can be formulated as:
 \begin{equation}
\begin{aligned}
    s_{E_{i}}&=S_{E_{i}}([E_{i}^{(L-1)}\cup e_{ic}^{(L)},q],M_{e})\\
\end{aligned}
\end{equation}
We select the top-K inferential evidence as current inferential evidence $E_{i}^{(L)}$ in the same way, through which achieving the revising and reconstructing of reasoning nodes. And, $E_{i}^{(L)}$ would be sent to Answerer as referenced evidence in the \textit{(L+1)-th} round.
\par It is important to note that since $E_{i}^{(L)}\in E_{s}^{(L)} \cup E_{i}^{(L-1)}$, and current quantities of information $E^{(L)}=E_{s}^{(L)} \cup E_{i}^{(L-1)}$, no new information is brought in after the updating of $E_{i}$. If Answerer fails to provide an effective answer with the knowledge context $E^{(L)}$, it is necessary to retrieve more source evidence to fill the knowledge gap. To achieve this, we should first know the information $E^{(L)}$ LLM already have right now, and the reason \textit{r} why LLM can`t (or wrongly) answer the question based on these information, then generate a new query to further retrieve information from the corpus to answer the initial question \textit{q}. Hence, we construct a LLM-based re-query generator $G_{R}$ with customized prompt $M_{R}$, to generate search query $q_{s}^{(L)}$ to deduce what information LLM needs to answer the question:   
 \begin{equation}
\begin{aligned}
    q_{s}^{(L)}&=LLM([E_{s}^{(L)},E_{i}^{(L)},r,q],M_{R} )\\
\end{aligned}
\end{equation}

\begin{table*}[!t]
\caption{Evaluation results on two multi-hop question answering datasets. '*' denotes the result outperforms baseline models in t-test at $p<0.05$ level. The best results are in \textbf{bold}, and the second best results are \underline{underlined}.}
\vspace{-3mm}
\centering
\begin{tabular}{ccccccccc}
\hline
 & \multicolumn{4}{c}{HotpotQA} & \multicolumn{4}{c}{2WikiMQA} \\ \cline{2-9}
\multirow{1}{*}{Methods} & EM & F1 & Pre& Rec& EM & F1 & Pre& Recall \\ \hline
Standard Prompting\cite{direct} & 14.2& 21.7& 23.2& 21.2& 21.4& 27.3& 28.8& 26.9\\
Chain-of-Thought\cite{COT} & 16.8& 24.9& 26.1& 24.9& 23.6& 30.4& 31.1& 30.5\\
Standard RAG\cite{rag_ori} & 23.4& 35.6& 36.6& 36.6& 22.8& 26.8& 27.2& 28.0\\
ReAct\cite{react} & 20.6& 29.4& 29.6& 32.1& 21.2& 28.5& 28.2& 30.3
\\
Self-Ask\cite{selfask} & 24.8& 35.1& 36.5& 36.4& 29.4& 36.7& 36.4& 38.2
\\
IR-COT\cite{ircot} & 30.4& 40.1& 41.6& 41.0& 25.6& 30.9& 31.0& 32.1\\
SearChain\cite{search-in-chain} & 29.6& 41.2& 41.5& 43.4& \underline{33.4}& \underline{42.6}& \underline{42.5}& \underline{44.8}\\
MetaRAG\cite{metarag} & \underline{32.4}& \underline{44.3}& \underline{45.5}& \underline{45.6}& 28.8
& 36.0& 35.7& 38.4\\
\textbf{RetroRAG} & \textbf{41.2}*& \textbf{54.9}*& \textbf{56.2}*& \textbf{58.3}*& \textbf{38.6}*& \textbf{46.6}*& \textbf{46.9}*& \textbf{49.5}*\\
\hline
\end{tabular}
\label{table:baselines}
\vspace{-2mm}
\end{table*}

\section{Experiments}
In this section, we evaluate the effectiveness of our proposed model on two multi-hop question answering (QA) datasets.
\subsection{Experimental Setup}
\subsubsection{Datasets and Evaluation Metrics}
We conduct experiments on two multi-hop question answering datasets: HotpotQA\cite{hotpotqa} and 2WikiMQA\cite{2wiki}. Since both of the datasets are constructed based on Wikipedia documents, we use the same document corpus and retrievers to provide external references for LLMs. Due to the constraints of experimental costs, following\cite{metarag}, we sub-sample 500 questions from the validation set of each dataset for experiments.
\par For evaluation metrics, we use exact match (EM) as our standard metrics at answer-level, to measure whether the predicted answer is completely consistent with the standard answer. And we use token-level F1, precision (Pre) and recall (Rec) for comprehensive evaluation at token-level, to evaluate the proportion of correct answer tokens in the overall tokens.   
\subsubsection{Baselines}
We compare our RetroRAG to recent baseline approaches: Standard Prompting\cite{direct}, Chain-of-Thought\cite{COT}, Standard RAG\cite{rag_ori}, ReAct\cite{react},  Self-Ask\cite{selfask}, IR-COT\cite{ircot}, SearChain\cite{search-in-chain}, and MetaRAG\cite{metarag}. We comprehensively describe each baseline in Appendix B.1 and explain the rationale behind selecting these specific baselines.
\subsubsection{Settings}
We choose GLM4-9B-chat\cite{chatglm4} LLM as the base LLM for all baseline and our RetroRAG approach with the temperature setting of 0.01, except the SC-generator of our RetroRAG whose temperature is set to 1.00. We utilize the Wikipedia dump\cite{wikipedia} as the document corpus for both datasets, where articles are segmented into passages of 100 tokens. We employ the BM25 algorithm\cite{bm25} and SimLM retriever\cite{simlm} to retrieve the top 5 relevant passages to be the external knowledge for all approaches. And we set a default judgment threshold for our answering evaluation mechanism at 0.7 to ensure consistency of answers. The maximum number of both iterations and the size of the evidence repository are set to 5. 

\subsection{Main Results}
Performance on multi-hop question answering datasets is shown in Table 1. It can be observed that:
\par (1) Our proposed RetroRAG consistently surpasses all baseline methods across two datasets. At answer-level, the performance improvement on EM is \textbf{+8.8} on HotpotQA and \textbf{+5.2}  on 2WikiMQA compared to the best baseline results; At token-level, the performance improvement on F1 is \textbf{+10.6} on HotpotQA and \textbf{+4.0} on 2WikiMQA compared to the best baseline results. This suggests that when directly employing LLM, without additional pre-training or fine-tuning, our approach exhibits optimal performance. 
\par (2) When compared to SearChain, which adapts the unidirectional forward reasoning paradigm but verifies each node in COT and outperforms other methods using the same paradigm, such as COT, ReACT, Self-Ask, etc., RetroRAG shows an improvement of \textbf{+11.6} on HotpotQA and \textbf{+5.2} on 2WikiMQA. This reflects that the retroactive reasoning paradigm RetroRAG uses can solve the issue of local insufficient reasoning, and provides more comprehensive reasoning, thereby improving the performance markedly.
\par (3) When compared to IR-COT and MetaRAG, which also do not adhere to the paradigm of linear reasoning, but increase the quantity of retrieved documents to re-generate the answer, RetroRAG shows an improvement of \textbf{+8.8} on HotpotQA and \textbf{+9.2} on 2WikiMQA. This reflects that the evidence-collation-and-discovery framework RetroRAG uses can address the issue of over-reasoning, and mitigate the irrelevant and noisy information from knowledge documents, thereby improving the performance significantly.
\par (4) Compared with Standard Prompting and COT approaches, both the idea of decomposing the initial query into multiple sub-questions and iteratively increasing the amount of knowledge retrieved can doubtlessly improve the ability of reasoning of LLMs. When multi-hop questions present a clearer progressive structure, such as samples in 2WikiMQA dataset, the approaches of decomposing sub-problems will perform better, so Self-Ask and SearChain demonstrate superior performance compared to IR-COT and MetaRAG in 2WikiMQA dataset since the latter may introduce excessive noise; On the contrary, the idea of collecting enough knowledge would help more in answering since the decomposing-answering mode could mislead the reasoning process, causes IR-COT and MetaRAG perform better on HotpotQA dataset. Due to our work's design of generation of inferential evidence and updating of evidence, which suppresses the introduction of noisy information and achieves a retroactive-progressive reasoning structure, achieving the best performance above all baselines on both datasets. 
\begin{table}[!h]
\caption{Ablation Studies on RetroRAG.}
\vspace{-3mm}
\centering
 \setlength{\tabcolsep}{4mm}{
\begin{tabular}{ccccc}
\hline
\multicolumn{1}{l}{} & \multicolumn{2}{c}{HotpotQA}         & \multicolumn{2}{c}{2WikiMQA}     \\ \cline{2-5} 
\multicolumn{1}{l}{} & EM        & F1              & EM        & F1              \\ \hline
\textbf{RetroRAG}                 & \textbf{41.2} & \textbf{54.9} & \textbf{38.6} & \textbf{46.6} \\ \hline
\multicolumn{5}{l}{Ablation of structure} \\ \hline
-w/o AE              & 38.6       & 51.7       & 32.6    & 40.1     \\ 
-w/o ELLERY              & 19.0       & 25.5       & 26.0     & 30.7     \\\hline
\multicolumn{5}{l}{Ablation of evidence utilization} \\ \hline
-w/o SE               & 34.8       & 46.1       & 29.2    & 35.6   \\
-w/o IE              & 38.4       & 51.9       & 36.6    & 44.5     \\ 
-w/o UoE               & 39.4       & 53.0      & 38.2     & 45.9  \\
-w/o EoE               & 40.5      & 54.8      & 38.0    & 45.8  \\ \hline
\end{tabular}
}
\vspace{-4mm}
\label{table:ablation}
\end{table}

\subsection{Ablation Study}
To verify the effectiveness of different components in RetroRAG, we conduct a comparative analysis respectively focus on frameworks, including Answerer (solely on the Answer Evaluation (AE) mechanism since Answerer needs to generate answer output) and ELLERY; Besides, we also focus on the designs of evidence utilization, including Source Evidence (SE), Inferential Evidence (IE), Updating of Evidence (UoE) and Evaluation of Evidence (EoE). The results are shown in Table 2.
\subsubsection{Ablation of structure} As shown in Table 2, it is evident that removing either of the frameworks adversely affects the performance of both datasets. Specifically, the removal of AE makes the LLM consistently set to a state of knowledge deficiency, thereby posing the risk of over-reasoning that turns the correct reasoning path built, into the incorrect one. Meanwhile, the removal of ELLERY would result in a complete absence of external knowledge, thereby leading to a significant decline in performance. This emphasizes the notion that many hallucinations stem from an insufficiency in external knowledge. 
\subsubsection{Ablation of evidence utilization} To conduct a more detailed analysis of the mechanisms of ELLERY, we performed ablation studies based on the characteristics of the evidence and the methods of evidence processing. Experimental results show that due to the complementary design, both source evidence and inferential evidence are crucial. Inferential evidence provides a summary of past effective information, thereby helping enhance the performance of LLMs. However, since it contains much less information than source evidence, source evidence has a greater impact on the quality of responses.  
Besides, the updating and evaluating of evidence can alleviate the introduction of irrelevant information, ensuring the progression of the current answering process while aligning with the initial question, thereby enhancing performance.

\subsection{Qualitative Analysis}

\begin{figure}[!t]
    \centering
  \includegraphics[width=\linewidth]{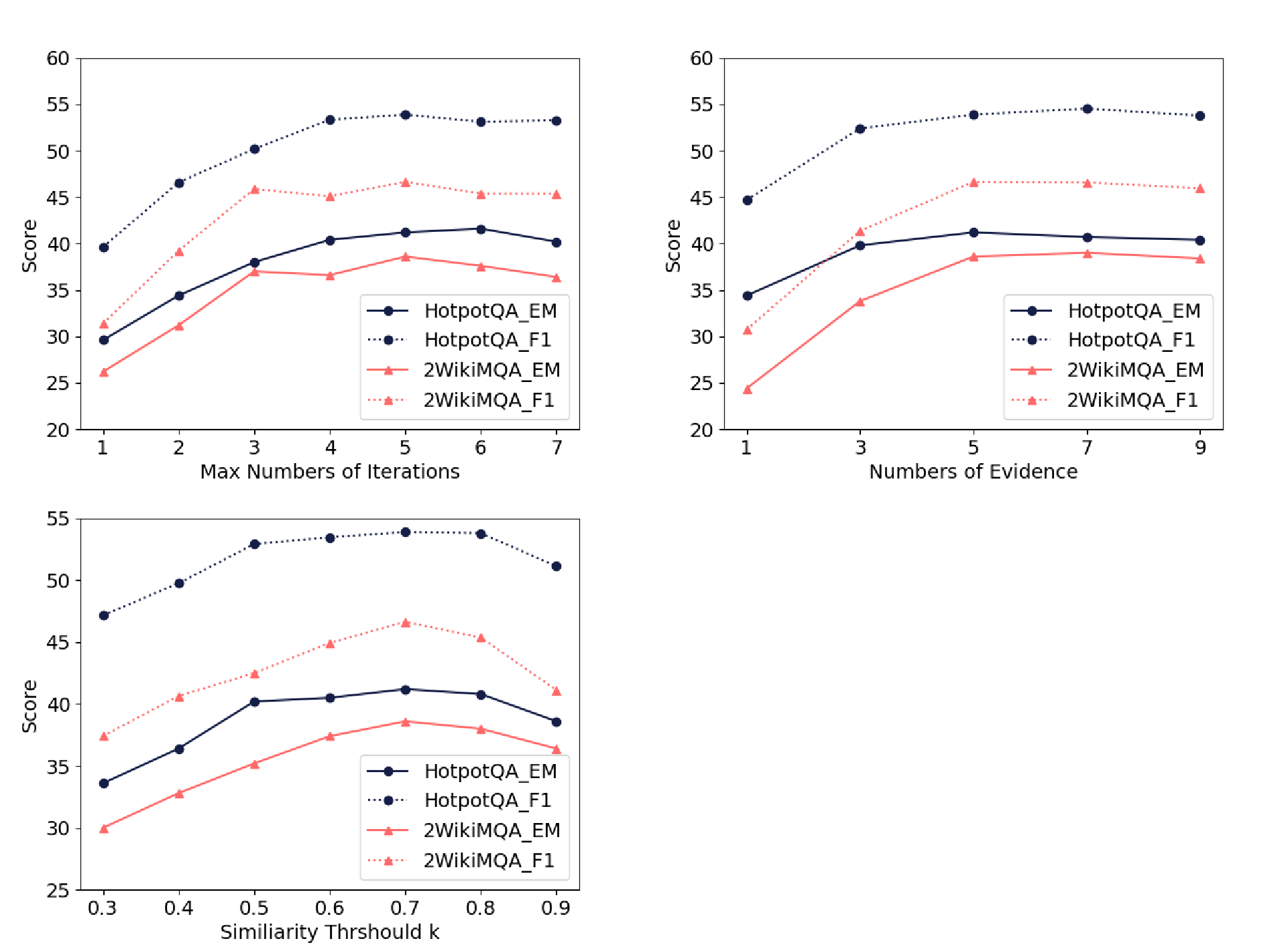}
  \vspace{-7mm}
  \caption{Comparison with different maximum numbers of iterations, numbers of evidence, and similarity thresholds.}
\label{fig:Comparison}
\vspace{-6mm}
\end{figure}

To better delve into the impact of the numbers of iterations and evidence, as well as similarity thresholds, we have embarked on a series of qualitative experiments. The results are shown in Figure 4. It can be observed that:
\subsubsection{Different maximum numbers of iterations} Although the Answer Evaluation mechanism plays a significant role in the performance of RetroRAG, the maximum number of iterations also has a substantial impact on the final results. As depicted in Figure 4, we can find that on both datasets, the accuracy of RetroRAG improves progressively before the iterations reach 3, and then grows relatively stable, peaking when the iterations reach 5 or 6. After the peak, we observed a risk of decline in the performance with the increase of iterations. We think these results suggest that, by increasing the number of iterations, RetroRAG can extract more effective evidence, thereby improving performance. However, excessively increasing may cause over-reasoning and introduce noise information, which in turn to a decline in performance. 
\subsubsection{Different numbers of evidence} For the convenience of the experiment, we set the numbers of both source evidence and inferential evidence to the same. Based on this setting, we design a series of experiments to investigate the impact of the number of evidence. As depicted in Figure 4, we find that on both datasets, increasing the numbers of evidence from 1 to 3 can significantly improve the performance of RetroRAG, which reflects that under the condition of insufficient information, it is difficult for LLMs to perform effective reasoning. The accuracy of RetroRAG reaches the peak at the number of 5 pieces of evidence, and then slowly decreases as the increase of the evidence. This suggests that although the filtering and updating mechanism can suppress irrelevant information to some extent, an excessively large evidence window can still introduce noise and impair performance. 
\subsubsection{Different similarity thresholds} 
Although threshold $t>0.5$ signifies that the LLM evaluator considers the result to be valid, a higher \textit{t} represents the LLM has more confidence in the determination. This could lead to a more reliable monitor but might also result in overly stringent requirements for the results, causing misjudgments and excessive iterations. As illustrated in Figure 4, we find that for HotpotQA dataset, $t=0.5$ can provide a good discriminative effect, but for 2WikiMQA, LLMs need a higher threshold which is about 0.7 to ensure the effectiveness. And for both datasets, excessively high thresholds lead to a decline in performance. This is also related to the over-reasoning issue.

\subsection{Case Study}

\begin{figure}[!t]
    \centering
  \includegraphics[width=\linewidth]{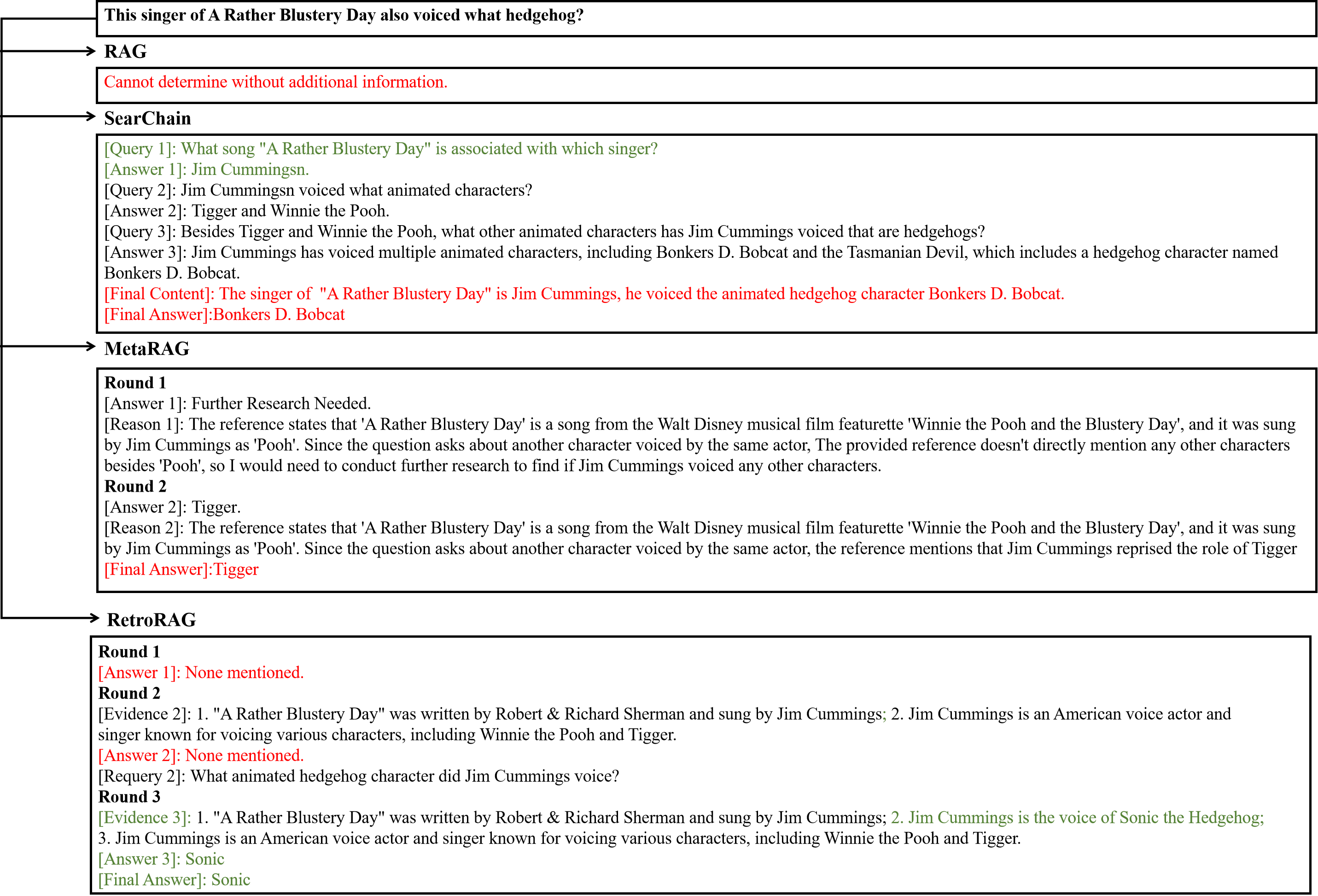}
  \vspace{-7mm}
  \caption{Case study of RetroRAG and previous approaches. }
\label{fig: case}
\vspace{-6mm}
\end{figure}

Due to factually related irrelevant documents from the inherent flaws of the current retrieval system,  previous approaches cause issues of insufficient reasoning and over-reasoning, respectively. To address these issues, we introduce inferential evidence as a form of knowledge caching. By summarizing and updating past relevant information, this approach enables retroactive awareness when new knowledge is introduced, while simultaneously reducing interference from irrelevant information. The case in Figure 5 provides an intuitive demonstration of the differences between our RetroRAG and previous approaches, traditional RAG stops reasoning due to insufficient retrieval information; SearChain makes the correct decomposing process but answers wrongly due to the related irrelevant knowledge; MetaRAG is interfered with excessive retrieval documents and makes the wrong reasoning process. Being aware of the evidence, RetroRAG makes the correct reasoning process that leads to an accurate answer. More cases would be given in Section B.2 of the Appendix.

\subsection{Effectiveness of evidence}
\begin{figure}[!t]
    \centering
  \includegraphics[width=\linewidth]{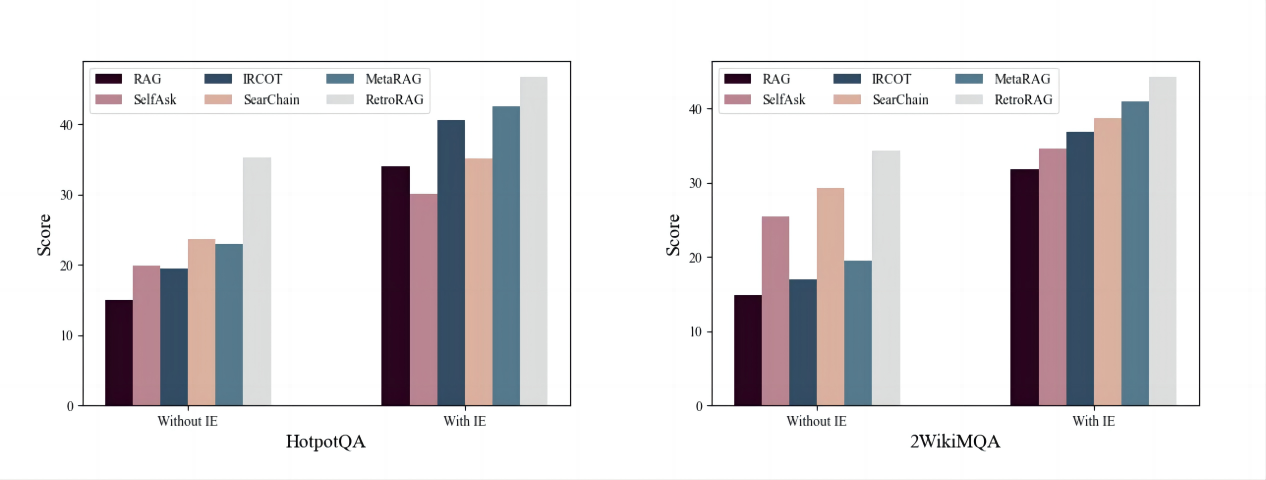}
  \vspace{-7mm}
  \caption{Comparison of different evidence cases of RetroRAG and previous approaches. }
\label{fig: case}
\vspace{-6mm}
\end{figure}
To better explore the effectiveness of Inferential Evidence, we divided the data into two categories based on the presence of evidence generated through Question-Relevance and Reference-Attribution evaluations during the iterative process. We then analyzed the differences in performance on the main baselines. As shown in Figure 6, we find in scenarios where Inferential Evidence is generated, iteratively increasing the amount of knowledge performs better compared to decomposing sub-questions. We believe this because the retrieval knowledge is extensively relevant and mutually corroborative, enhancing the LLM's performance. Furthermore, our work enhances the LLM's understanding of knowledge by generating Inferential Evidence, thereby further improving the reliability of its responses. Additionally, the absence of Inferential Evidence generation may be due to the sparse distribution of required knowledge within individual documents. In such cases, decomposing sub-questions allows the LLM to better understand and answer the questions. Moreover, our work ensures the accumulation of effective information through Source Evidence updating, thereby also enhancing the LLM's performance.

\section{Conclusion}
In this paper, we point out the threat from the unidirectional forward reasoning paradigm inherent in traditional RAG methods, within which any errors produced during reasoning steps are irreversible and affect the whole reasoning chain. We then introduce RetroRAG, a novel framework that uses a detective-like retroactive reasoning paradigm that can revise and reconstruct the reasoning chain, ensuring it on the correct direction. Through the evidence-collation-discovery framework, RetroRAG can search, generate, and update credible evidence, empower the model to perceive existing information, and seek out more necessary evidence to complete the reasoning process. Experimental results on two multi-hop QA datasets demonstrated that RetroRAG performs better than all baselines. In the future, we aspire to explore the possibility of allowing LLMs to independently learn the aforementioned evidence-collation-discovery process through methods such as fine-tuning or pre-training.

\bibliography{aaai25}

\clearpage
\appendix
\section{A. Prompt Detail}
We show the prompt used in experiment on both datasets in Figure 7 to Figure 13. In this work, we constructed the few-shot CoT prompt, and the declarative assessor prompt by referencing \cite{metarag}, and design different prompt for the corresponding functions. For all calculations as detailed in the Methodology section, we quantified the results by generating probability distributions of the 'yes' and 'no' tokens.

\begin{figure}[H]
    \centering
  \includegraphics[width=\linewidth]{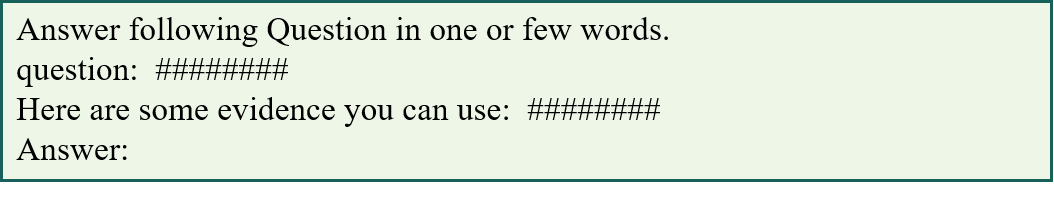}
  \vspace{-7mm}
  \caption{Prompt $M_{D}$ for generating the monitoring answer.}
\label{fig: prompt-SC}
\vspace{-6mm}
\end{figure}

\begin{figure}[H]
    \centering
  \includegraphics[width=\linewidth]{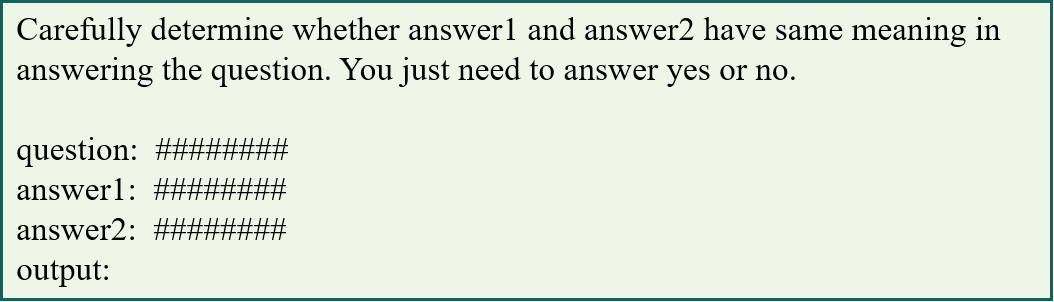}
  \vspace{-7mm}
  \caption{Prompt $M_{sc}$ for calculating the Self-Consistency score to determine if the answer is reliable.}
\label{fig: prompt-SC}
\vspace{-6mm}
\end{figure}

\begin{figure}[H]
    \centering
  \includegraphics[width=\linewidth]{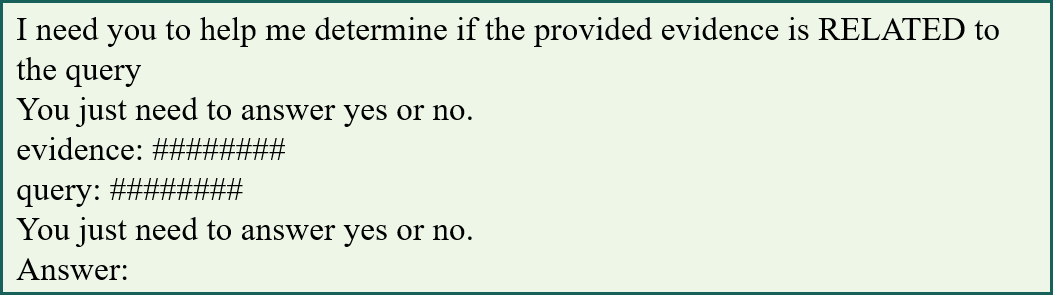}
  \vspace{-7mm}
  \caption{Prompt $M_{e}$ for calculating the score of relevance between evidence and query for evidence updating.}
\label{fig: prompt-rerank}
\vspace{-6mm}
\end{figure}

\begin{figure}[H]
    \centering
  \includegraphics[width=\linewidth]{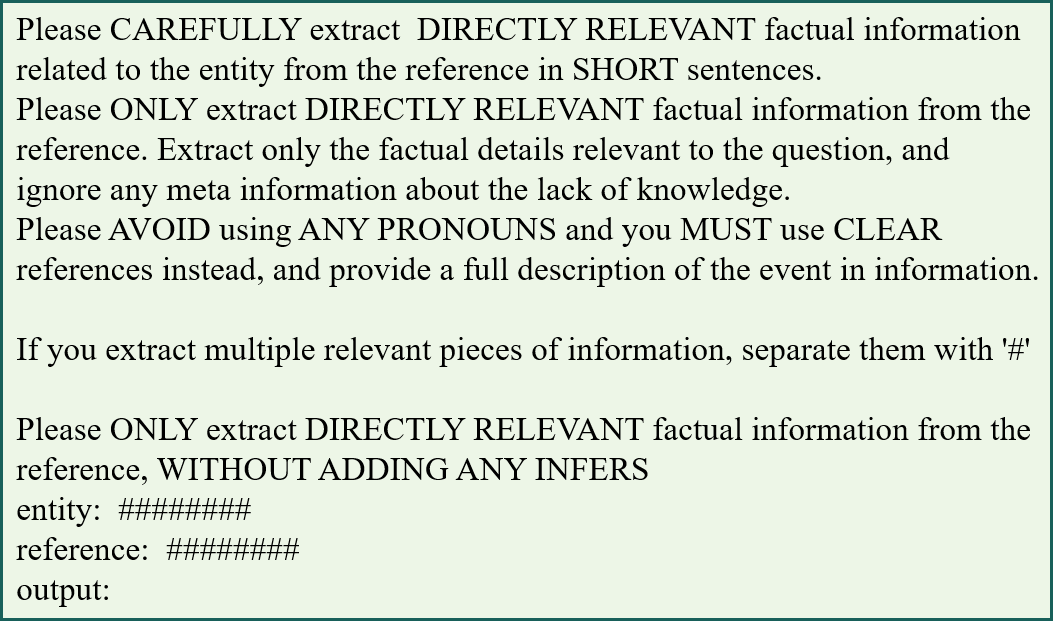}
  \vspace{-7mm}
  \caption{Prompt $M_{IE}$ for generating inferential evidence.}
\label{fig: prompt-evidence_generate}
\vspace{-6mm}
\end{figure}

\begin{figure}[H]
    \centering
  \includegraphics[width=\linewidth]{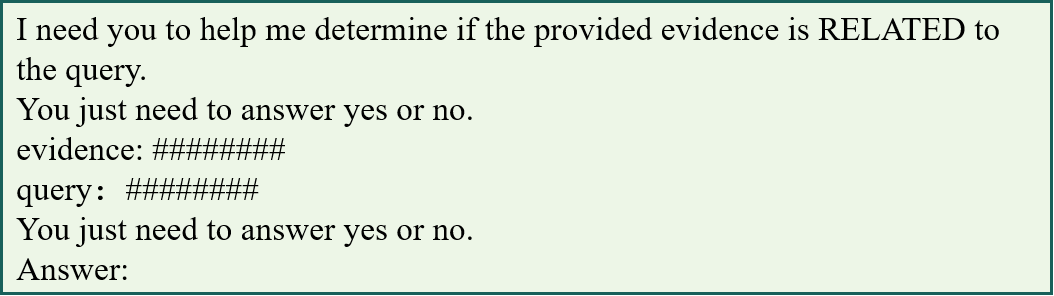}
  \vspace{-7mm}
  \caption{Prompt $M_{qr}$ for calculating the Question-Relevance score.}
\label{fig: prompt-Evidence_related}
\vspace{-6mm}
\end{figure}

\begin{figure}[H]
    \centering
  \includegraphics[width=\linewidth]{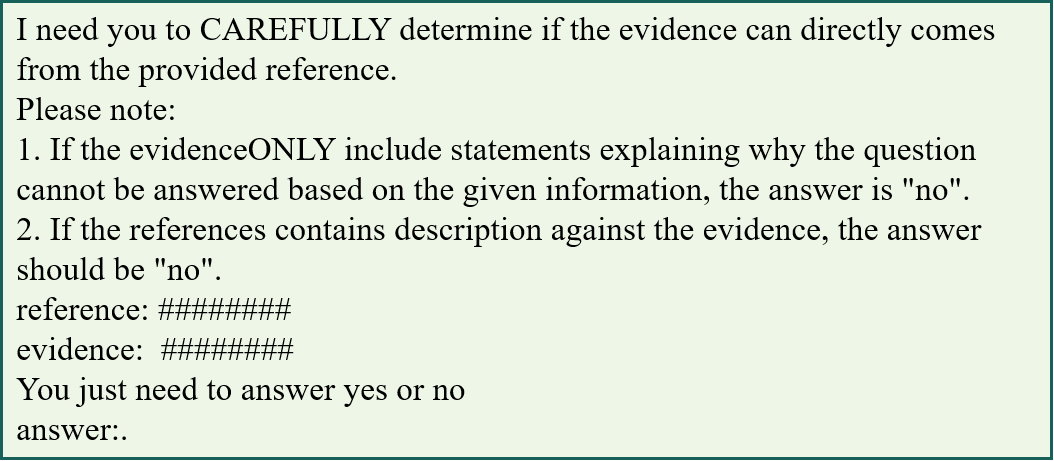}
  \vspace{-7mm}
  \caption{Prompt $M_{ra}$ for calculating the Reference-Attribution score.}
\label{fig: prompt-RA}
\vspace{-6mm}
\end{figure}

\begin{figure}[H]
    \centering
  \includegraphics[width=\linewidth]{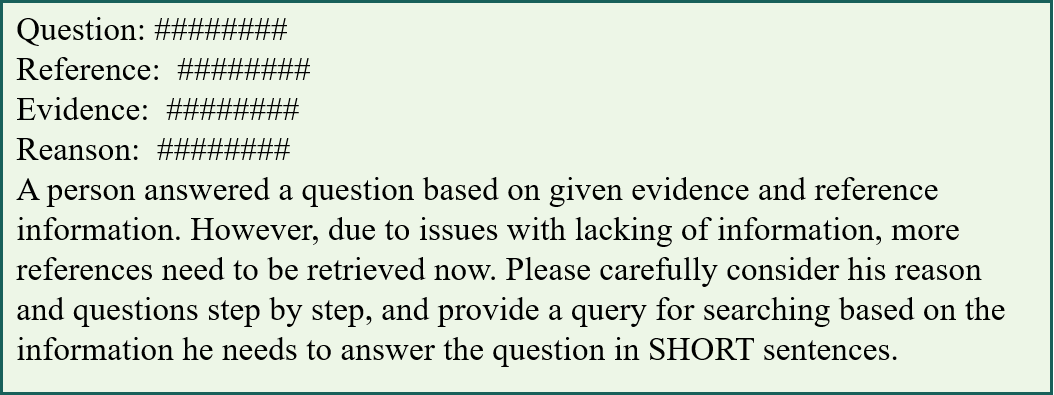}
  \vspace{-7mm}
  \caption{Prompt $M_{R}$ for generating re-query.}
\label{fig: prompt-requery}
\vspace{-6mm}
\end{figure}

\section{B. Experimental Details}
\subsection{Baselines} We compare our proposed model with several state-of-theart baselines listed as follows:
\begin{itemize}
    \item Standard Prompting\cite{direct}: Standard Prompting directs LLM to answer the queries with a simple question-answer prompt.
    \item Chain-of-Thought\cite{COT}: Chain-of-Thought provides a few-shot prompt to LLM, make it can answer with a reasoning process.
    \item Standard RAG\cite{rag_ori}: Standard RAG first retrieves multiple documents by query, then inject these documents into prompts to LLM for answering.
    \item ReAct\cite{react}: ReAct introduces a reasoning and acting paradigm, alternately executing reasoning and task-specific actions to complete the QA task. 
    \item Self-Ask\cite{selfask}:  Self-Ask introduces a paradigm that decomposes questions into sub-questions, continuously engaging in self-questioning until the final answer is obtained. 
    \item IR-COT\cite{ircot}: IR-COT iteratively alternates between using COT reasoning to guide retrieval, and utilizing retrieval results to enhance CoT reasoning, continuing executing until the final answer is obtained. 
    \item SearChain\cite{search-in-chain}: SearChain introduces the concept of "search-in-chain," which corrects the reasoning process through the interaction between Information Retrieval (IR) and Chain-of-Query (COQ), let IR porvides the knowledge that LLM really needs.
    \item MetaRAG\cite{metarag}: MetaRAG combines the RAG process with metacognition, allowing LLM to execute different actions based on the reliability of internal and external knowledge, identify the sufficiency of knowledge and potential errors during reasoning. 
\end{itemize}
\subsection{Additional Case Studies}
We present additional cases to further demonstrate the effectiveness of our proposed RetroRAG approach. As shown in Figure 14, we find that RetroRAG not only leverages the effective information from inferential evidence to make accurate reasoning, but can also compensates for insufficient inferential evidence by retrieving complementary source evidence through re-query, thereby enabling correct reasoning, which also highlights the necessity of leveraging both inferential evidence and source evidence. 
\begin{figure}[H]
    \centering
  \includegraphics[width=\linewidth]{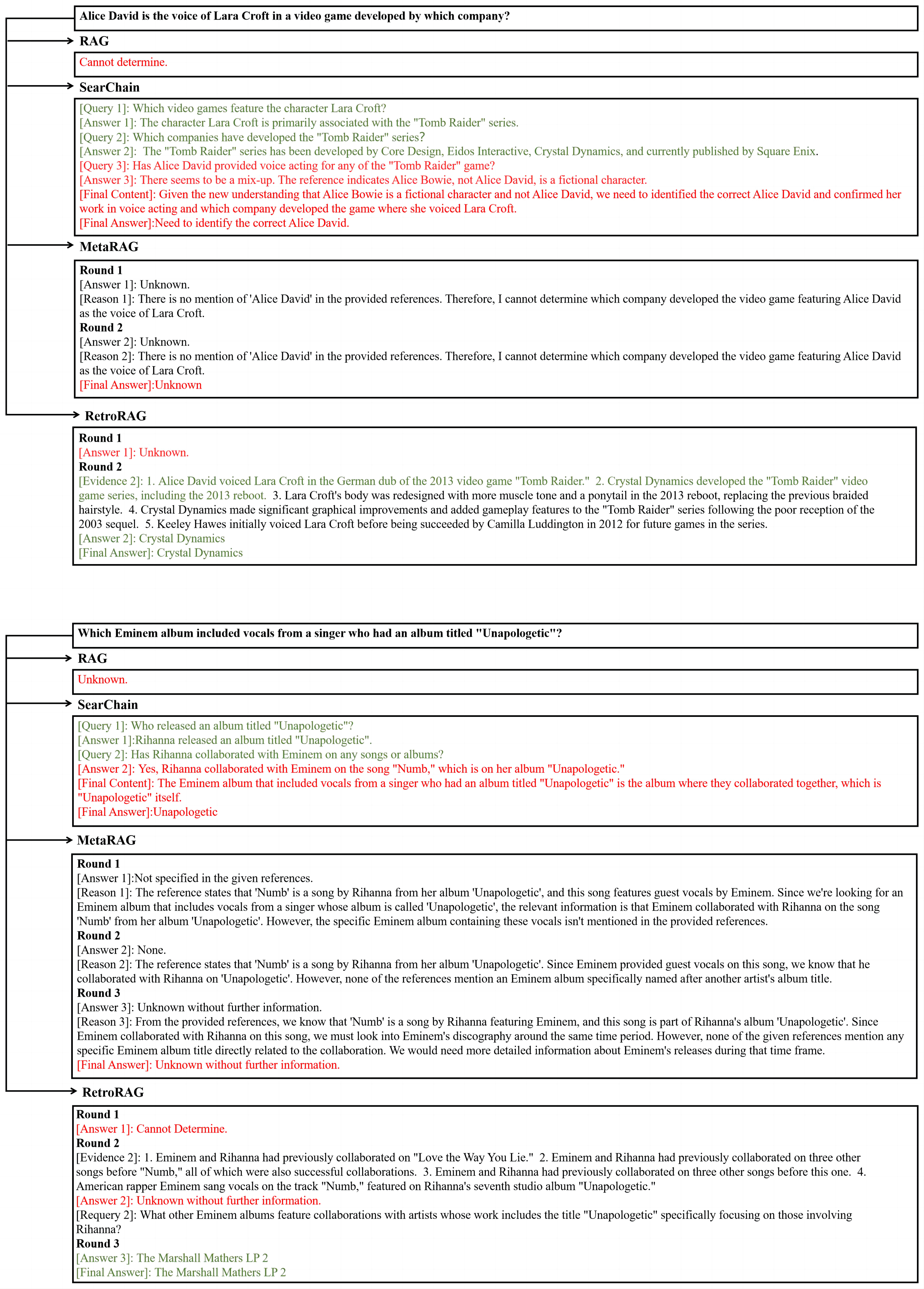}
  \vspace{-7mm}
  \caption{Additional Case Studies.}
\label{fig: casestudy}
\vspace{-6mm}
\end{figure}

\section{C. Assessment using LLM-judge across Various Scenarios    }
To validate the transferability of our approach across different linguistic contexts, and its reliability in simple question-answering scenarios, we conducted performance evaluations on the knowledge question-answering segments of a Chinese hallucination evaluation dataset HalluQA\cite{HalluQA}, which contain single-hop question build from Baidu Baike. We utilize the Baidu Baike dump as the document corpus for HalluQA datasets, where articles are segmented into passages of 100 tokens, and we employ the BM25 algorithm to retrieve the top 5 relevant documents to be the external knowledg.Given that the golden truth in HalluQA contains detailed descriptions, making it challenging to quantify performance using token-level metrics, we employed the GPT4 as the LLM-Judge, with the same setting of \cite{HalluQA}, to assess the answer semantic accuracy. 
\par Additionally, we simultaneously applied the same LLM-Judge to evaluate the performance on HotpotQA and 2WikiMQA dataset from the semantic perspective. 
\begin{table}[!h]
\caption{LLM-Judge on three datasets.}
\vspace{-3mm}
\centering
 \setlength{\tabcolsep}{2mm}{
\begin{tabular}{cccc}
\hline
\multicolumn{1}{l}{} & \multicolumn{1}{c}{HotpotQA}         & \multicolumn{1}{c}{2WikiMQA}  & \multicolumn{1}{c}{HalluQA}   \\ \cline{2-4} 
\multicolumn{1}{l}{} & LLM-Judge        & LLM-Judge               & LLM-Judge                  \\ \hline
Standard              & 29.4       & 29.2       & 27.7     \\ 
COT              & 32.0       & 32.4       & 37.0      \\
RAG               & 47.4       & 38.6       & 45.2  \\
ReAct              & 42.3       & 35.3       &36.6 \\
Self-Ask             & 52.0       & 39.7      &30.4  \\
IR-COT             & 53.4       & 42.5      & 56.8   \\
SearChain             & 56.8       & 47.9      & 48.3   \\
MetaRAG             & 56.2       & 40.9      & 57.8   \\
RetroRAG               & 68.2      & 51.2      & 64.3   \\ \hline
\end{tabular}
}
\vspace{-4mm}
\label{table:ablation}
\end{table}
\par As shown in Table 3, our RetroRAG outperforms all baselines in semantic perspective on all three datasets, This indicates that our approach demonstrates more stable and reliable performance across various scenarios. Also, we observed that some baselines with unidirectional forward reasoning paradigm, exhibit performance degradation when encountering simple questions with single-hop.  For instance, Self-Ask may degrade to standard prompting, ReAct may degrade to CoT, and SearChain may degrade to RAG. This highlights once again the necessaries of constructing a retroactive reasoning process.
\end{document}